\documentclass[11pt]{article}
\usepackage{acl2015}
\usepackage{times}
\usepackage{url}
\usepackage{color}
\usepackage[utf8]{inputenc}
\usepackage{graphicx}
\usepackage{latexsym}
\usepackage{tabularx}
\usepackage[dvipsnames]{xcolor}
 \usepackage{graphicx}
 \usepackage[numbers]{natbib}
 \usepackage{hyperref}
 \usepackage{mhchem, siunitx, makecell}
\usepackage{multirow}
\usepackage{subfig}
\usepackage{hyperref}

\title{Zero-Shot Language Transfer vs Iterative Back Translation for Unsupervised Machine Translation }

\author{ Aviral Joshi, Chengzhi Huang, Har Simrat Singh\\
  {\tt (aviralj, chengzhh, harsimrs)@andrew.cmu.edu}
}

\date{}

\begin{document}
\maketitle
\begin{abstract}
This work focuses on comparing different solutions for machine translation on low resource language pairs, namely, with zero-shot transfer learning and unsupervised machine translation. We discuss how the data size affects the performance of both unsupervised MT and transfer learning. Additionally we also look at how the domain of the data affects the result of unsupervised MT. The code to all the experiments performed in this project are accessible on \href{https://github.com/chanzy3/11747_Final_Project}{Github}.
\end{abstract}

\section{Introduction}
Unsupervised Machine translation(UMT) is the task of translating sentences from a source language to a target language without the help of parallel data for training. UMT is especially useful when translating to and from low resource languages for example English to Nepali or vice-versa. Since it is assumed that there is no access to parallel data between two languages, unsupervised MT techniques focus on leveraging monolingual data to learn a translation model. More generally there are three data availability scenarios that can dictate the choice of the training procedure and the translation model to be used. These scenarios are as follows:

\begin{enumerate}
    \item We have no parallel data for \textbf{X-Z} but have a large monolingual corpus. Unsupervised Machine translation is very well suited for this scenario.  
    \item We do not have parallel data for \textbf{X-Z} but we have parallel data for \textbf{Y-Z} and a good amount of monolingual data. Zero-Shot language transfer can be more effective in this scenario.   
    \item Finally, we have some parallel data for \textbf{X-Z} and monolingual data for \textbf{X} and \textbf{Z}. In this scenario semi-supervised approaches that leverage both monolingual and parallel data can outperform the 2 other approaches mentioned.
\end{enumerate}
Here \textbf{X} represents the source language (the language to transfer from), \textbf{Z} represents the target language (the language to transfer to) and \textbf{Y} represents a pivot language (an intermediate language) which is similar to the target language \textbf{Z}. In a low resource scenario there is little to no parallel data between \textbf{X} and \textbf{Z}.

The first 2 data availability scenarios mentioned above are generally more challenging due to lack of presence of parallel data between source and target languages and are explored in greater detail in this work. Futhermore, our report also discuss the impact of mismatch in dataset domains for training and evaluation and its effect on translation.
Our report is structured as follows. We first introduce related work in the domain of unsupervised machine translation with back translation and language transfer while also discussing works that detail the impact of domain mismatch in training and evaluation datasets for UMT in Section \ref{sec:related}. After which we describe the experimental setup in Section \ref{sec:experiments}. Moving on we describe the results of our experiments on low-resource language pairs with varying training data sizes and also analyze the impact of out of domain datasets on the quality of translation in Section \ref{sec:results}. Finally, we conclude with possible directions for future work in the domain of UMT for low-resource languages. 

\section{Related Work}
\label{sec:related}

\subsection{Unsupervised Machine Translation Using Monolingual Data}
Recent advancement in Machine Translation can be attributed to the development of training techniques that allow deep learning models to utilize large-scale monolingual data for improving performance on translation while reducing dependency on massive amounts of parallel data. Of the several attempts to leverage monolingual data three techniques -- back-translation, language modelling and Auto-Encoding -- in particular standout as the most effective and widely accepted strategies to do so. 

Sennrich et. al. \cite{sennrich2016improving}  first introduced the idea of iteratively improving the quality of a machine translation model in a semi-supervised setting by using “back-translation”, the idea revolved around training an initial sub-optimal translation system with the help of available parallel data and then utilizing it to translate many sentences from a large monolingual corpus on the target side to the source language to generate augmented parallel-data for training the original translation system.

Gulcehre et. al \cite{gulcehre2015using} demonstrated an effective strategy to integrate language modelling as an auxiliary task with Machine Translation improving the performance on translation quality between both high resource languages and low resource language pairs. A more recent work by Lample. et. al \cite{lample2019crosslingual} introduced a Cross Lingual Model(XLM) that utilized a Cross Lingual Language modelling pretraining objective to improve performance on translation.

Previous works have also looked at the task of translation from the perspective of an Auto-Encoder to utilize monolingual data. Cheng et. al.\cite{cheng2019semi} introduced an approach which utilized an Auto-Encoder to reconstruct sentences in monolingual data where the encoder translates input from source to target language and the decoder works to translate sentences back to the source language. Parallel data is utilized in conjunction to monolingual data to learn the translation system. This method was shown to outperform the original back-translation system proposed by Sennrich et. al.

Unlike the previously described semi-supervised Lample et. al. \cite{lample2017unsupervised} introduce an effective strategy to perform completely unsupervised machine translation in the presence of a large monolingual corpus. Their work leverages a Denoising Auto-Encoder (DAE) to reconstruct denoised sentences from the sentences in monolingual corpora which are systematically corrupted to train encoder and decoders for both source and target language. To map the latent DAE representations of the two languages into the same space. To aid translation the authors also introduce an additional adversarial loss term. The DAE along with iterative back-translation is utilized to train a completely unsupervised translation system.

\subsection{Transfer Learning - Zero shot Translation}
Even though UMT has shown to perform well on similar language pairs with good amount of monolingual data,  \citeauthor{artetxe2020rigor} argued a scenario without any parallel data and abundant monolingual data is unrealistic in practice. One common alternative in this scenario is to avoid using monolingual data with zero-shot transfer learning. Zero-shot translation concerns the case where direct (src-tgt) parallel data is lacking but there is parallel data between the target language and a similar language to source language.

Transfer Learning is firstly proposed for NMT by \citeauthor{zoph-etal-2016-transfer}, who leverage a high resourced parent model to initialize the low-resourced child model. \citeauthor{nguyen2017transfer} and \citeauthor{Kocmi_2018} utilizes shared vocabularies for source and target language to improve transfer learning, in addition, mitigate the vocabulary mismatch by using cross-lingual word embedding. These methods show great performance in transfer learning for low resource languages, they show limited effects on zero-shot transfer learning.

Multilinguality has been extensively studied and has been applied to the related problem of zero-shot translation (\citeauthor{johnson-etal-2017-googles}; \citeauthor{firat2016zeroresource}; \citeauthor{arivazhagan2019missing}). \citeauthor{liu2020multilingual} showed some initial results on multilingual unsupervised translation in the low-resource setting. Namely, Pretrained mBART25 is finetuned on transfer language to target language pair, and is directly to the source to target pairs. mBART obtained up to 34.1 (Nl-En) with zero-shot translation and works well when fine-tuning is also conducted in the same language family. Specifically, it acheives 23 BLEU points for Ro-En using Cs as the transfer language, and 18.9 BLUE points for Ne-En using Hi as the transfer language.

\subsection{Unsupervised Machine Translation on Out-of-Domain Datasets}
Owing to increasing popularity of unsupervised MT and the techniques involved, adaptation of UMT to translate data from domains where a large amount of parallel data is not obtainable is also gaining importance. One of the main issues associated with this is the feasibility of using pre-trained models. Most of the pre-trained models, even though trained under supervised learning paradigm, fail to achieve significant translation accuracy on a domain which was non-existent in the training data. Unsupervised models also follow the same pattern. \citeauthor{marchisio2020does} \cite{marchisio2020does} show some significant results and comment on the domain similarity of test and training data in State of the Art models. The authors use United Nations Parallel Corpus (UN) \cite{ziemski-etal-2016-united}, News Crawl and CommonCrawl datasets for their work. The model was trained solely on NewsCrawl while being evaluated on all 3 datasets. A general trend of reduction in performance was observed across various language pairs. Extensive work has been done to address this problem. \citeauthor{dou2019unsupervised} \cite{dou2019unsupervised} propose a Domain-Aware feature embeddings based approach and achieve improvements over previous existing methods. \citeauthor{luong2015stanford} \cite{luong2015stanford} adopt an attention based approach to fine tune a pretrained model on out of domain data. \citeauthor{sennrich2016improving} \cite{sennrich2016improving} show that for out of domain data back translation helps the models learn a whole lot of words previously non existent in the vocabulary which could be the reason why out of domain back translation based UMT fails. Our experiments follow a similar line of work and show that fine tuning a pre-trained translation model does not effectively help with out of domain data. 

\section{Experiments}
\label{sec:experiments}

\subsection{Proposed Experiments}
In our experiments, we focus on how the size of training data will affect the performance of both zero-shot transfer learning and Unsupervised Machine Translation with iterative back translation. Additionally, we conduct an extensive empirical evaluation of unsupervised machine translation using dissimilar domains for monolingual training data between source and target languages and with diverse datasets between training data and test data. 

\begin{enumerate}
    \item \textbf{Unsupervised MT using iterative back translation}: Our first experiment involves UMT using back translation(BT) with the XLM model. Here we demonstrate the feasibility of UMT with  BT for low-resource language pairs and also how variation in the monolingual data size effects the translation quality for similar and dissimilar language pairs.  
    
    \item  \textbf{Zero Shot Language Transfer}: Our second experiement involves zero-shot translation using mBART model. We finetune on related language (\textbf{X} $\rightarrow$ En) and apply it directly to the source-target (\textbf{Y} $\rightarrow$ En) language pair. Here we illustrate how the size of finetuning data affect both finetuning scores (finetuned on \textbf{X} $\rightarrow$ En and tested on \textbf{X} $\rightarrow$ En) and transferring scores (finetuned on \textbf{X} $\rightarrow$ En and tested on \textbf{Y} $\rightarrow$ En). 
    
    \item  \textbf{Domain Dissimilarity}
    Our third experiment analyses the performance of a model which is trained on a different dataset and is tested on sentences from an entirely different domain. We use XML model and perform back translation on pre-trained models for all language paris \textbf{X} $\rightarrow$ \textbf{En} to fine tune it on the required dataset. Back translation steps include \textbf{X} $\rightarrow$ \textbf{En} and \textbf{En} $\rightarrow$ \textbf{X}. The model is then tested for the previously mentioned steps on a dataset sourced from a completely different domain.
\end{enumerate}

\subsection{Models}

mBART \cite{liu2020multilingual} is used for experiments with zero-shot transfer learning, 
XLM \cite{lample2019crosslingual} is used for experiments with unsupervised machine translation.

\begin{enumerate}
    \item  
    mBART is the first method for pre-training a complete sequence to sequence model by denoising full texts in multiple languages. Pre-training is done on CC25 data set and pretrained model can be directly fine-tuned for both supervised and unsupervised MT. It also enables a smooth zero-shot transfer to language pairs with no bi-text data. In our experiments, we use mBART25, which is pretrained on all 25 lagnauges. Ideally, for a fair comparison unsupervised BT should be also done using mBART25, however, the authors did not provide their implementation for BT and so we shift to the XLM model for experiments with BT since its implementation was more accessible.
    \item XLM is a cross lingual language model developed for both supervised and unsupervised machine translation tasks. XLM model uses shared sub word vocabulary and is trained on 3 different language modelling objectives of Causal language modeling, Masked Language modeling and Translation language modeling. The model is essentially Transformer based with 1024 sub units and 8 heads. Cross lingual XLM is trained on 15 languages and is shown to achive SoTA on all the different languages. Our motivation behind using XLM was that a pretrained XLM model was available for the languages of our choice. It supported 2 way backtranslation and was comparatively easier to train and fine tune than most other multilingual models.
\end{enumerate}

\subsection{Dataset}

Two language pairs were chosen for the experiments where we conduct an investigation of the impact of size of monolingual training data on translation quality between a similar and dissimilar language pair (Ro (Romanian) - En (English) and Ne (Nepali) - En respectively. For Ro \& En, data from wikipedia was used. We perform random sampling to obtain monolingual sentences from the monolingual data provided by the authors of the XLM model (This was originally obtained by scraping articles from Wikipedia). For Nepali the monolingual data was obtained from the Flores data corpus.

For zero-shot transfer learning, we conducted experiments on Ro  to En with Cs (Czech) being the transfer language and Ne to En with Hi (Hindi) being the transfer language. Cs - En parallel training data is taken from Europarl v7 \cite{koehn2005epc} with 668,595 sentence pairs, Ro - En test data is taken from WMT16 test set for Romanian and English, consisting of 1999 sentence pairs; Hi-En parallel training data is taken from IITB Bombay English-Hindi corpus\cite{DBLP:journals/corr/abs-1710-02855} with 1,609,682 sentence pairs, Ne-En test data is taken from \cite{flores} with 2835 test sentences.


To analyse the effect of domain mismatch training and testing, we use Fr (French), Genrman (De) and Romanian (Ro). In all the experiments the target language is English (En). Back translation is performed both ways. The choice for these specific languages was motivated by the criterion on language similarity and availability of data. French and German are related to English, albeit not closely but still possess a similarity in terms of language family and closeness of vocabulary. Romanian was chosen as a low resource language, distantly related to English. To simulate domain mismatch, the XLM model was pretrained on WMT'16 which mostly has parallel data extracted from news articles. The model was then fine tuned on monolingual data of the four languages obtained from the Tatoeba challenge \cite{tiedemann2020tatoeba}. For French, English and German 50K monolingual sentences were used, and for Romanian only 1999 sentences were available. Tatoeba dataset has sentences with minimum overlap with any other available machine translation dataset. We used the sentences extracted from Wikipedia pages present within the dataset. The validation and testing set was extracted from WMT'16. 

\subsubsection{Experiment Details}
We use the following tokenizers for both preprocessing training data and evaluations for all three experiments:
\begin{itemize}
    \item \textbf{Ne, Hi}: We use Indic-NLP Library to tokenize both Indic language inputs and outputs.\cite{kunchukuttan2020indicnlp}
    
    \item \textbf{Ro, Cs, En}: We apply Moses tokenization and special normalization for English, Czech and Romanian texts. 
\end{itemize}

 We conduct all the experiments for transfer learning on AWS \verb|g4dn.xlarge| instance, with \verb|max-tokens| 512, \verb|total-num-update| 80000 and dropout 0.3. During the inference time, beam size is set to 5.
 
Similar to experiments with transfer learning all experiments for Back Translation were conducted on a g4dn.xlarge instance the batch size was set to 32 and multiple iterations of back translation were run. The number of tokens per batch were set to a maximum of 1500 and a embedding dimension of 1024 was used. Other hyper-parameters were kept the same as the implementation provided by the authors of XLM.
 
 For domain mismatch experiment we use XLM model as well. We follow the Byte Pair Encoding technique as used by the authors of XLM, and use fastBPE for the same. Tokenization is done using Moses tokenization. The experiments were conducted on AWS \verb|p3.2xlarge| instances. The model used 6 layers of transformers with 8 heads. The embedding dimension was 1024.
 
\subsection{Evaluation Metrics}

For all of our tasks, we use BLEU score calculated by \verb|sacrebleu| as the automatic metric to evaluate the translation performance. Specifically, we compute the BLEU scores over tokenized text for both hypothesis outputs and references as mentioned in \citeauthor{liu2020multilingual}

\section{Results and Discussions}
\label{sec:results}

In this section we describe the results of the proposed experiments as mentioned previously in Section \ref{sec:experiments} and discuss our finding to illustrate the the impact of dataset size and dataset domain for Unsupervised Machine Translation and data size for Transfer Learning.   

\subsection{Back Translation}
As expected with most deep learning approaches, the trained model scales with increase in training data (monolingual) as the BLEU show a general upward trend as more data is made accessible to the XLM model. This is evident in the results shown in Table \ref{tab:umt_bt} especially for the high similarity English-Romanian language pair. Whereas, the performance on low similarity English-Nepali pair is degenerate in all data availability scenarios. Even when the entire training corpus is provided to the model. This result indicates that it might not be feasible to perform unsupervised MT when language pairs have little or no similarity using Back Translation along. In which case other approaches for Unsupervised MT such as language transfer are better suited. More about language transfer is discussed in Section \ref{subsec:transfer_learning}.

A big improvement in BLEU score is seen when training data size is increased from 10,000 to 100,000 for the high similarity language pair. The performance increases steadily as more data is provided to the model while the rate of increase slows down. A reasonable BLEU for the English-Nepali language pair was reached with just 600,000 sentences from each language.

\begin{table}[h!]
    \begin{tabular}{|c|r|r||r|r|}
     \hline
     \multicolumn{5}{|c|}{BLEU - Back Translation} \\
     \hline
     \multicolumn{1}{|c}{Data Size} & \multicolumn{2}{|c||}{High Similarity} & \multicolumn{2}{|c|}{Low Similarity} \\
     \hline
      & en & ro & en & ne \\
      \hline
     & $\rightarrow$  & $\leftarrow$ & $\rightarrow$ & $\leftarrow$ \\
     \hline
     \hline
     10K & 2.76 & 3.93 & 0.0 & 0.0\\
     100K & 23.6 & 23.6 & 0.0 & 0.0\\
     300K & 27.7 & 26.9 & 0.1 & 0.1\\
     600K & 28.8 & 28.3 & 0.1 & 0.2\\
     \textbf{$>$10M} & \textbf{33.3}  & \textbf{31.8} & \textbf{0.1} & \textbf{0.5}\\
     
    \hline
 \end{tabular}
 \caption{Comparison of Translation Quality between Similar and Dissimilar language pairs and how they scale with increase in monolingual data.}
 \label{tab:umt_bt}
\end{table}

\begin{figure*}
    \centering
    \subfloat[]{{
        \includegraphics[width=8.3cm]{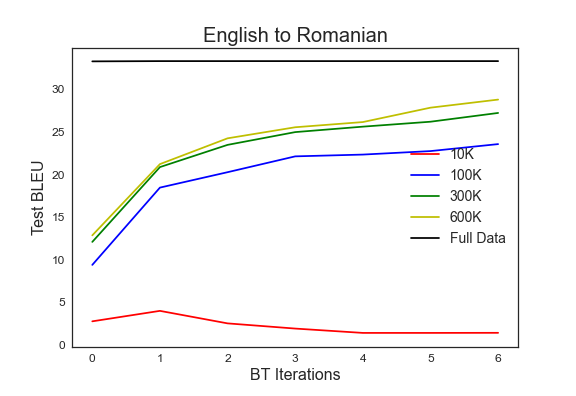}
    }}
    \subfloat[]{{
        \includegraphics[width=8.3cm]{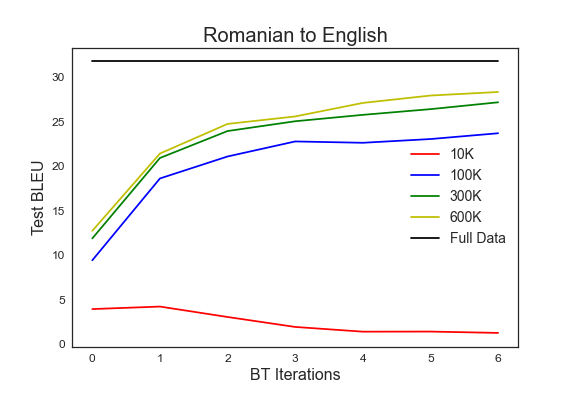}
    }}
    \caption{Figures depicting the variation of translation quality with the availability of monolingual data for English and Romanian}
\end{figure*}

\subsection{Transfer Learning}
\label{subsec:transfer_learning}
Zero shot transfer learning for mBART \cite{liu2020multilingual} works as following: pretrained mBART is finetuned on a related language \textbf{Y} to target En and is directly applied for testing on source \textbf{X} to target language En. 

In order to investigate on how the size of training data will affect the zero-shot BLEU points, we randomly sample 1K, 10K, 100K and 300K sentences from Europart v7 \cite{koehn2005epc} of Cs - En parallel training corpus and test the finetuning model directly on Ro - En test data; We also randomly sample 10K, 100K, 300K and 600K sentences from ITTB Bombay-English corpus \cite{DBLP:journals/corr/abs-1710-02855} and test the finetuning model directly on Ne - En test data. We present BLEU points of finetuning models on both source-target language pairs (\textbf{X} - En) and related-target language pairs (\textbf{Y} - En) for better observing the correlation between them. Each experiment is trained using the same hyper-parameters.
\begin{table}[h!]
    \begin{tabular}{|c|c|c|}
     \hline

    \textbf{Data Size} & \thead{\textbf{Transfer} \\ \textbf{Ro} $\rightarrow$ \textbf{En}} & \thead{\textbf{Finetune} \\ \textbf{Cs} $\rightarrow$ \textbf{En}}  \\
     \hline
     0 &  0.0  & 0.0 \\
     1K & 17.6 & 16.2 \\
     10K & \textbf{23.3} & 21.9 \\
     100K & 23.0 & 25.7\\
     300K & 20.3 & 26.5\\
     668K & 14.5 & \textbf{27.5}\\
     
    \hline
 \end{tabular}
 \caption{BLEU points for Transfer Learning and directly finetuning of different data size, pretrained mBART25 is finetuned on \textbf{Cs} to \textbf{En} and directly transferred to \textbf{Ro} to \textbf{En}, we choose the best checkpoint weights validated on \textbf{Cs} to \textbf{En} dev set}
 \label{tab:tt1}
\end{table}

\begin{table}[h!]
    \begin{tabular}{|c|c|c|}
     \hline

    \textbf{Data Size} & \thead{\textbf{Transfer} \\ \textbf{Ne} $\rightarrow$ \textbf{En}} & \thead{\textbf{Finetune} \\ \textbf{Hi} $\rightarrow$ \textbf{En}}  \\
     \hline
     0   &  0.0 & 0.0 \\
      10K & 11.7 & 12.8 \\
     100K & 14.0 &  18.7\\
     300K & 13.6 & 19.1\\
     600K & 13.2 & 19.3\\
     1.68M & \textbf{15.1} & \textbf{23.4} \\
     
    \hline
 \end{tabular}
 \caption{BLEU points for Transfer Learning and directly finetuning of different data size, pretrained mBART25 is finetuned on \textbf{Hi} to \textbf{En} and directly transferred to \textbf{Ne} to \textbf{En}, we choose the best checkpoint weights validated on \textbf{Hi} to \textbf{En} dev set}
 \label{tab:tt2}
\end{table}

It is worth noting that for the model finetuned on Cs - En, we observed a generally negative correlation between transferring scores and finetuning scores, as well as a negative correlation between the data size and transferring scores. In contrast, for the model finetuned on Hi - En there is a positive correlation between transferring scores and finetuning scores and a positive correlation between the data size and transferring scores. One hypothesis is that even though mBART achieves the highest transferring scores for Ro - En when finetuned on Cs - En and they both belong to Indo-European language family, they are still vastly different languages using orthographic distance metric \citeauthor{8c2cca6c565646e8a15e745aa9571d41} developed, with Czech being in the Slavic Language sub-group and Romanian being in the Romance language sub-group. Hence, finetuning on a larger data set would make mBART overfit towards the finetuned language, thus hurts the performance of zero-shot translation on source-target pair due to the limited capacity of mBART model itself. For Ne and Hi are more closely related to each other and have large overlap in their vocabularies, so finetuning on Hi will greatly benefit the performance of zero-shot translation.

\subsection{Comparison between Back Translation with Zero-Shot Transfer Learning}

\textbf{When does back translation work? } In Table \ref{tab:tt2}, we do not have much increase in performance when we increase the data size for monolingual data when training on dissimilar language pairs (Ne and En). If there exists high quality similar languages bi-text data, language transfer is able to beat the conventional methods with BT. For similar language pairs (Ro and En), we do see the gain in BLEU points when data size increases and can easily beat the zero-shot transfer learning with 100K monolingual data.

\textbf{When does transfer learning work?} Note that in Table \ref{tab:tt1}, the performance for zero-shot transfer learning suffers from "the curse of data size", which means simply increasing the size of finetuning data does not contribute to the performance of transferring scores, especially when finetuning language (Cs) is not close enough to the target language (Ro). Romanian is closest to Catalan \cite{8c2cca6c565646e8a15e745aa9571d41}, however, not so much bi-text exists between Catalan and English and it is also a out of vocabulary language for mBART, thus does not benefit much from multilingual pretraining. On the contrary, for languages in Indic Language family, increase in data size does help in both transferring scores and direct finetuning scores. Compared with Back Translation, zero-shot transfer learning outperforms it with just a small amount of bi-text for Hi to En.

\subsection{Domain Dissimilarity}
\begin{table}[h!]
    \begin{tabular}{|c|c|c|}
     \hline
     \multirow{2}{*}{Language Pair} & \multicolumn{2}{c|}{Training Dataset} \\ \cline{2-3} 
                               & WMT'16      & Tatoeba(Wikipedia)      \\ \hline
    \hline
     Fr $\rightarrow$ En & 32.8 & 27.38 \\
     De $\rightarrow$ En & 34.3 & 27.44 \\
     Ro $\rightarrow$ En & 18.3 &  1.74 \\
     \hline
     En $\rightarrow$ Fr & 33.4 & 28.33 \\
     En $\rightarrow$ De & 26.4 & 21.93 \\
     En $\rightarrow$ Ro & 18.9 &  0.39 \\
     \hline
 \end{tabular}
 \caption{Domain Mismatch}
 \label{tab:dom_bt}
\end{table}

We present the results of domain mismatch experiments in table \ref{tab:dom_bt}. As expected there is a consistent decrease in the performance of the model across all languages. The decrease is comparatively less for Fr and De then Ro. French and German are borh high resource languages and the XLM model for both these languages was pre trained as well as finetuned on 50K sentences. Whereas for Romanian only 1999 sentences were available in the Tatoeba challenge. The size of the validation and test data was also smaller. Using back translation in this scenario did not prove to be ideal. This also suggests that even for high resource languages the domain of the dataset is a big factor to determine the overall translation efficacy of the model. XLM uses shared sub word vocabulary which means there would be a lot of words which are differently used in different domains and are thus harder for the model to predict. The results for Romanian could have been better if transfer learning was used as the UMT technique. The results are particularly interesting as for low resource languages like Romanian, getting parallel domain specific data is very tough and adapting a pre-trained model does not help in  the same way as it does in cases of high resource languages. We wish to explore this further with some extra experiments as well.

\section{Conclusion and Future-Work}
In this work, we compare two different solutions for low resource language MT: unsupervised MT with BT and zero-shot transfer learning. We focus on how the size affects the performance for both unsupervised MT and transfer and how the domain matters in unsupervised MT.

For transfer learning, we find that if the model is transferred from a "not so similar" language, like Cs $\rightarrow$ En in our experiment, the performance for transfer learning can not be scaled up by simply increasing the size of finetuning data. In contrast, for the closely related language, like languages in Indic Language Family, we observe an opposite trend. However, the capacity of mBART model has not yet been well investigated, and we still do not have a clear definition of the similarity between transfer language and target language. These could be the potential topics conducted in the future resesarch.

Talking about domain mismatch, we wish to explore if transfer learning is a better UMT technique for adapting a model to a different domain in case of low resource languages. We also want to try more experiments to further solidify and generalize our findings. There have been numerous researches done in this field where attempts have been made to learn domain specific embeddings or even model based changes to generalize across mismatching domains. We hope to use some of the existing methods and try to boost the performance of models specifically in case of low resource languages. 

\bibliography{main}
\end{document}